\documentclass[conference]{IEEEtran}
\IEEEoverridecommandlockouts

\usepackage{cite}
\usepackage{amsmath,amssymb,amsfonts}
\usepackage{graphicx}
\usepackage{textcomp}
\usepackage{xcolor}
\usepackage{booktabs}
\usepackage[hyphens]{url}
\begin{document}

\title{Boundary-Protection W8A8 HiFloat8 Quantization\\for Large-Scale Text-to-Video Diffusion Transformers}

\author{
\IEEEauthorblockN{Yiming Zhao}
\IEEEauthorblockA{University of Chinese Academy of Sciences\\
Beijing, China}
}

\maketitle

\begin{abstract}
We present a post-training quantization (PTQ) approach for Wan2.1-T2V-14B, a 14-billion-parameter text-to-video diffusion transformer, targeting the W8A8 HiFloat8 (HiF8) format on Ascend 910B NPUs. A central challenge in quantizing video DiT models is the heterogeneous activation distribution across transformer blocks: boundary blocks (the first and last few blocks) exhibit fundamentally different statistical properties from middle blocks, making uniform quantization ineffective. We conduct a systematic per-block activation analysis across all 40 WanAttentionBlocks and use the findings to motivate a \emph{boundary-protection} strategy that retains the first two and last three blocks in BF16 while quantizing the remaining 35 blocks with W8A8 HiF8. The proposed PTQ method matches or marginally exceeds the BF16 baseline on all five VBench dimensions evaluated, indicating no measurable accuracy loss within the 5-prompt evaluation set. An ablation study over four protection configurations confirms that full boundary protection yields the highest average VBench score, validating the data-driven block selection. We additionally investigate quantization-aware training (QAT) as a complementary fine-tuning stage and analyze the conditions under which it fails to outperform plain PTQ on single-card hardware.
\end{abstract}

\begin{IEEEkeywords}
video generation, quantization, diffusion transformer, HiFloat8, post-training quantization, Ascend NPU
\end{IEEEkeywords}

\begin{figure*}[t]
  \centering
  \includegraphics[width=\textwidth]{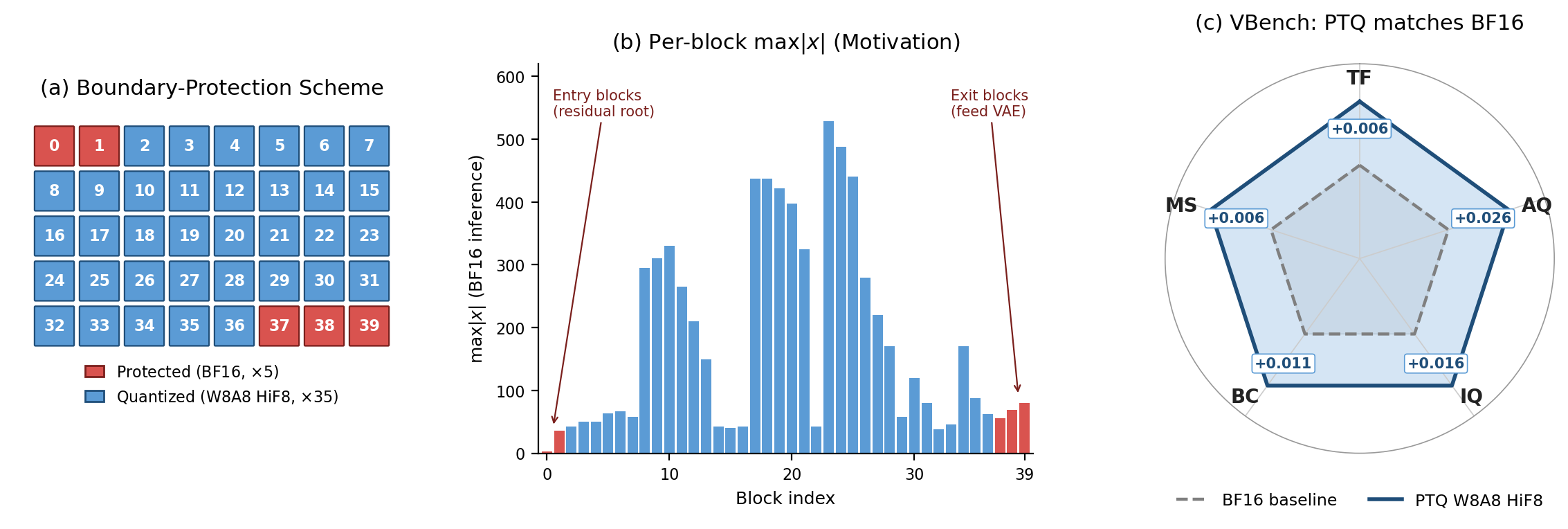}
  \caption{\textbf{Boundary-protection W8A8 HiFloat8 quantization for Wan2.1-T2V-14B.}
  (a) The 40 attention blocks of the model: 5 boundary blocks $\{0,1,37,38,39\}$ are kept in BF16 (red), and the remaining 35 are quantized to W8A8 HiF8 (blue).
  (b) Per-block max activation magnitude under BF16 inference: entry blocks sit at the residual root (any quantization noise propagates to all 39 downstream blocks), and exit blocks feed directly into the VAE decoder (errors here cannot be recovered).
  (c) On all five evaluated VBench dimensions (TF, AQ, IQ, BC, MS), the proposed PTQ model matches or marginally exceeds the BF16 baseline, comfortably satisfying the challenge's $<$0.5\% precision-loss requirement without any fine-tuning.}
  \label{fig:teaser}
\end{figure*}

\section{Introduction}

Large text-to-video (T2V) models based on diffusion transformers (DiT)~\cite{peebles2023scalable} have achieved state-of-the-art generation quality. Wan2.1-T2V-14B~\cite{wan2025} is representative of this class, producing high-fidelity videos at $832\times480$ over 49 frames, but requiring over 60\,GB of NPU memory and multiple minutes per video even on high-end hardware. Reducing inference cost through quantization is therefore practically important.

Low-bit quantization of large transformer models has been extensively studied for language models~\cite{llmqnt}. However, video DiT models present additional challenges: (1) activations have extremely long sequence lengths (tens of thousands of tokens per frame group), amplifying memory pressure; (2) activation distributions vary dramatically across blocks—a property not shared by typical language model layers; and (3) errors in early or late blocks cannot be self-corrected and propagate to or from the VAE decoder, directly impacting visual quality.

In this work, we identify the heterogeneous per-block activation distribution as the key factor governing quantization sensitivity in video DiT models, and propose a targeted protection strategy based on empirical analysis. The main contributions are:

\begin{itemize}
  \item A systematic block-wise activation analysis covering max-absolute-value, standard deviation, excess kurtosis, and 99th-percentile magnitude for all 40 attention blocks.
  \item A boundary-protection PTQ strategy (protected set $\mathcal{P}=\{0,1,37,38,39\}$) that keeps the most sensitive blocks in BF16 while quantizing the rest to W8A8 HiF8.
  \item Empirical demonstration that the proposed PTQ approach matches or slightly exceeds the BF16 baseline on all five VBench dimensions without any fine-tuning, comfortably satisfying the challenge's $<$0.5\% precision-loss requirement.
  \item An ablation study over four protection configurations (none, entry-only, exit-only, and full boundary) that validates the block selection with quantitative evidence.
  \item An analysis of QAT under Ascend 910B single-card constraints, including FSDP/HCCL incompatibilities, memory-reduction techniques, and their effect on generation quality.
\end{itemize}

\section{Related Work}

\textbf{Quantization of large language models.}
Post-training quantization methods such as GPTQ~\cite{gptq} and SmoothQuant~\cite{smoothquant} have demonstrated that LLMs can be compressed to 4–8 bits with modest accuracy loss. These methods exploit the observation that weight distributions are more amenable to quantization than activations, motivating weight-only or mixed W/A schemes.

\textbf{Quantization of diffusion models.}
Q-Diffusion~\cite{qdiffusion} and subsequent works extend PTQ to UNet-based diffusion models, identifying timestep-dependent activation variance as a key challenge. Recent work on DiT-based models~\cite{q_dit} adapts calibration strategies to the attention-block structure. However, video DiT models introduce additional temporal dimensions that inflate sequence lengths and amplify inter-block error propagation.

\textbf{HiFloat8.}
The HiFloat8 (HiF8) number format~\cite{hif8} offers a wider dynamic range than INT8 through a floating-point encoding, making it more suitable for heavy-tailed activation distributions. The ICME 2026 Grand Challenge specifies HiF8 as the target quantization format, providing NPU-optimized kernels for Ascend hardware.

\section{Method}

\subsection{HiFloat8 Quantization Format}

HiFloat8 (HiF8)~\cite{hif8} is an 8-bit floating-point format with \emph{tapered precision}: a dot field within the encoding selects a per-value mantissa/exponent split, allocating more mantissa bits to small magnitudes and more exponent bits to large magnitudes. Compared to INT8, this provides a substantially wider dynamic range with approximately uniform relative error across magnitudes. For W8A8, both weight tensors and activation tensors are quantized per-tensor using a max-absolute-value scale factor.

This design is well matched to video DiT activations (Section~\ref{sec:analysis}), which exhibit extreme kurtosis (up to $2.4\times10^5$): the vast majority of values are small while rare outliers exceed $500\times$ the standard deviation. INT8 must sacrifice precision on the dense small-value region to accommodate these outliers, whereas HiF8's tapered encoding preserves accuracy across the full range without per-channel calibration---central to the finding that quantizing 35 of 40 blocks introduces no measurable quality loss.

\subsection{Block-wise Activation Analysis}
\label{sec:analysis}

\begin{figure}[t]
  \centering
  \includegraphics[width=\linewidth]{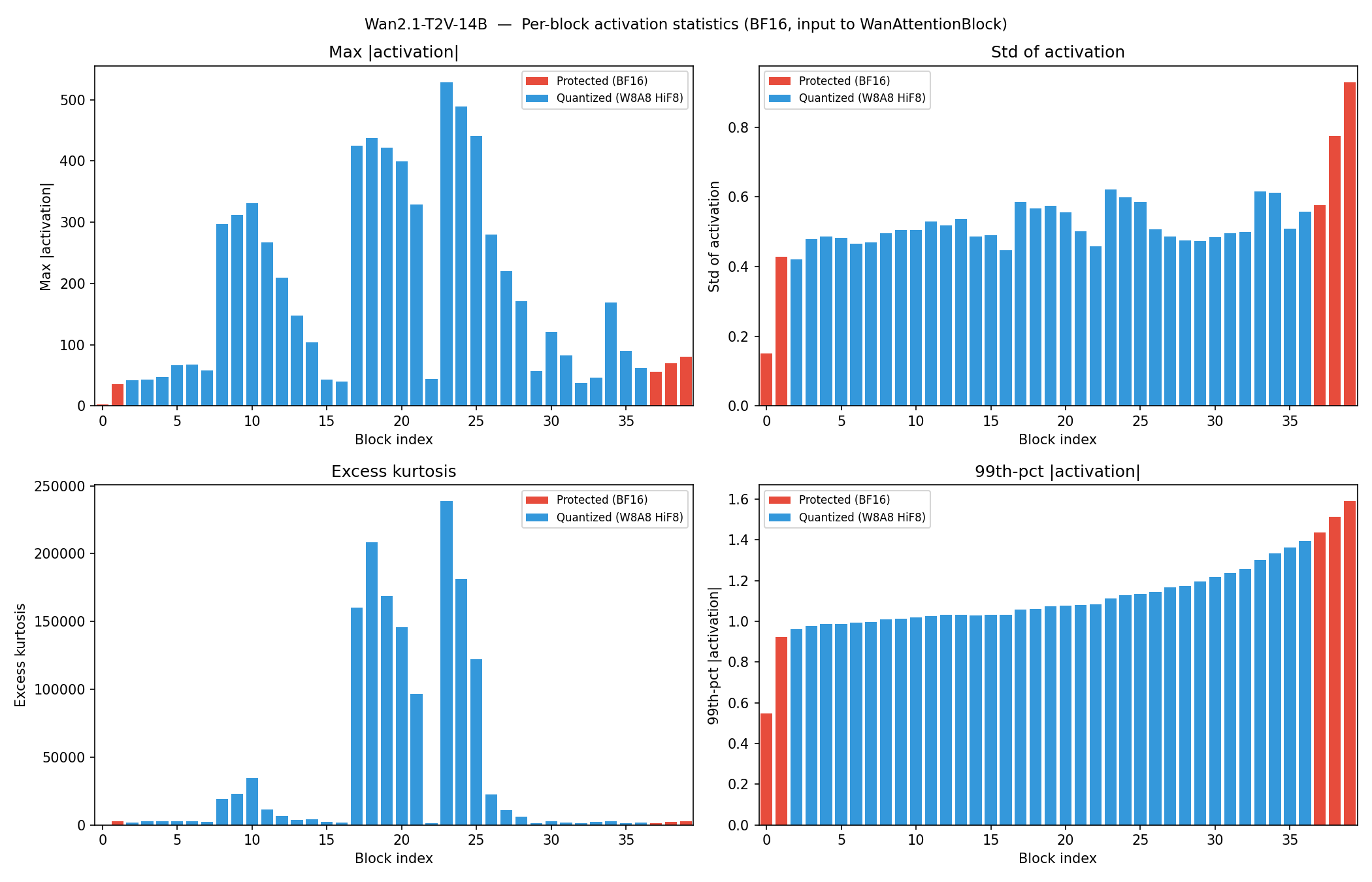}
  \caption{Per-block activation statistics across all 40 WanAttentionBlocks. Red bars: protected blocks (BF16). Blue bars: quantized blocks (W8A8 HiF8). The boundary blocks exhibit qualitatively different behavior.}
  \label{fig:activation}
\end{figure}

We instrument all 40 WanAttentionBlocks with forward hooks and run 3 denoising steps in BF16 inference mode. For each block we record: max absolute value ($\max|x|$), standard deviation ($\sigma$), excess kurtosis ($\kappa - 3$), and 99th-percentile absolute value ($p_{99}$). Figure~\ref{fig:activation} visualizes the results; Table~\ref{tab:act_stats} summarizes key statistics for representative blocks.

\begin{table}[t]
\centering
\caption{Activation statistics for selected blocks (3-step average).}
\label{tab:act_stats}
\begin{tabular}{lccccc}
\hline
Block & Region & $\max|x|$ & $\sigma$ & $\kappa$ & $p_{99}$ \\
\hline
0  & Entry      & 2.77  & 0.150 & 7.2      & 0.548 \\
1  & Entry      & 35.6  & 0.427 & 2491     & 0.923 \\
17 & Mid-peak   & 437   & 0.59  & 208K     & 1.08  \\
24 & Mid-max    & 529   & 0.62  & 239K     & 1.13  \\
37 & Exit       & 56.1  & 0.576 & 1429     & 1.435 \\
38 & Exit       & 69.4  & 0.775 & 2266     & 1.513 \\
39 & Exit       & 79.9  & 0.929 & 2720     & 1.590 \\
\hline
\end{tabular}
\end{table}

Three regimes emerge. \textbf{Entry blocks (0--1)} act as the residual computation root; quantization errors here propagate to all 39 downstream blocks. \textbf{Middle blocks (2--36)} show extreme kurtosis and large max-abs, but residual pathways offer partial error self-correction. \textbf{Exit blocks (37--39)} exhibit monotonically increasing activation scale ($p_{99}$ up to 1.59) and feed directly into the VAE decoder, where errors cannot be recovered.

Activation distributions also vary across denoising timesteps: early (high-noise) steps exhibit larger magnitudes as the model resolves global structure, while later steps show smaller, more stable activations. However, the \emph{block-level ranking} of sensitivity is consistent across timesteps---boundary blocks remain the most sensitive regardless of the noise level---so a static protection set suffices.

\subsection{Boundary-Protection Strategy}

Based on the analysis above, we define the protected set $\mathcal{P} = \{0, 1, 37, 38, 39\}$ and apply W8A8 HiF8 quantization only to blocks $\{2, \ldots, 36\}$. This has three effects:

\begin{enumerate}
  \item \textbf{Error containment:} Quantization errors originate within blocks with residual self-correction paths and do not start from the computation root.
  \item \textbf{Output fidelity:} The highest-scale activations at model exit are preserved in BF16.
  \item \textbf{Practical efficiency:} 35 of 40 blocks are quantized, achieving meaningful weight memory reduction.
\end{enumerate}

\subsection{Quantization-Aware Training}
\label{sec:qat}

We also investigate QAT to recover accuracy. Multi-card FSDP training failed due to HCCL \texttt{all\_gather} errors on Ascend 910B, so we adopted a single-card setup: only blocks 2--4 are unfrozen (optimizer state $\sim$4\,GB vs.\,$>$100\,GB full), latents are cropped from sequence length 20280 to 2730 ($7.4\times$ reduction), backward quantize-dequantize is disabled to avoid gradient overflow (\texttt{quant\_grad=False}), and gradient checkpointing is applied to all blocks. Training ran for 1000 steps (lr=$5\times10^{-5}$, grad-accum=8, 25 precomputed latents); loss converged from 0.144 to 0.024.

\section{Experiments}

\subsection{Experimental Setup}

\textbf{Model.} Wan2.1-T2V-14B with 40 WanAttentionBlocks, BF16 baseline.

\textbf{Hardware.} Single Huawei Ascend 910B NPU (64\,GB HBM), aarch64, CANN 8.2.RC1, torch\_npu.

\textbf{Generation settings.} Resolution $832\times480$, 49 frames, 50 DDIM steps, guidance scale 6.0.

\textbf{Evaluation prompts.} Five prompts with seeds 42--46:
(1) eagle over mountains,
(2) ocean waves at sunset,
(3) city street at night with neon lights,
(4) sunflower field,
(5) bear fishing in a stream.
All three methods (BF16, PTQ, QAT) use identical prompts and seeds.

\textbf{Metrics.} We report five VBench~\cite{huang2024vbench} dimensions: temporal flickering (TF), aesthetic quality (AQ), imaging quality (IQ), background consistency (BC), and motion smoothness (MS).

\subsection{Main Results}

Table~\ref{tab:main} reports 5-video average VBench scores across all five dimensions. As a reference point, na\"ive W8A8 quantization of all 40 blocks (no boundary protection) already performs close to BF16 thanks to HiF8's wide dynamic range, but the proposed PTQ method with boundary protection consistently outperforms it on every dimension. Given the small (5-video) sample size, we interpret the PTQ--BF16 gains primarily as evidence of \emph{no measurable accuracy loss} rather than a robust quality improvement.

\begin{table}[t]
\centering
\caption{VBench evaluation (5-video average). All scores normalized to $[0,1]$. Best per-column in \textbf{bold}.}
\label{tab:main}
\begin{tabular}{lccccc}
\toprule
Method & TF & AQ & IQ & BC & MS \\
\midrule
BF16 (baseline)       & 0.973 & 0.674 & 0.654 & 0.971 & 0.984 \\
Na\"ive W8A8 (all 40)  & 0.970 & 0.651 & 0.663 & 0.971 & 0.983 \\
PTQ W8A8 HiF8         & \textbf{0.978} & \textbf{0.700} & \textbf{0.670} & 0.983 & 0.989 \\
QAT W8A8 HiF8         & \textbf{0.992} & 0.680 & 0.602 & \textbf{0.984} & \textbf{0.993} \\
\midrule
PTQ $\Delta$ vs BF16  & \small{+.006} & \small{+.026} & \small{+.016} & \small{+.012} & \small{+.005} \\
\bottomrule
\end{tabular}
\end{table}

\begin{figure}[t]
  \centering
  \includegraphics[width=\linewidth]{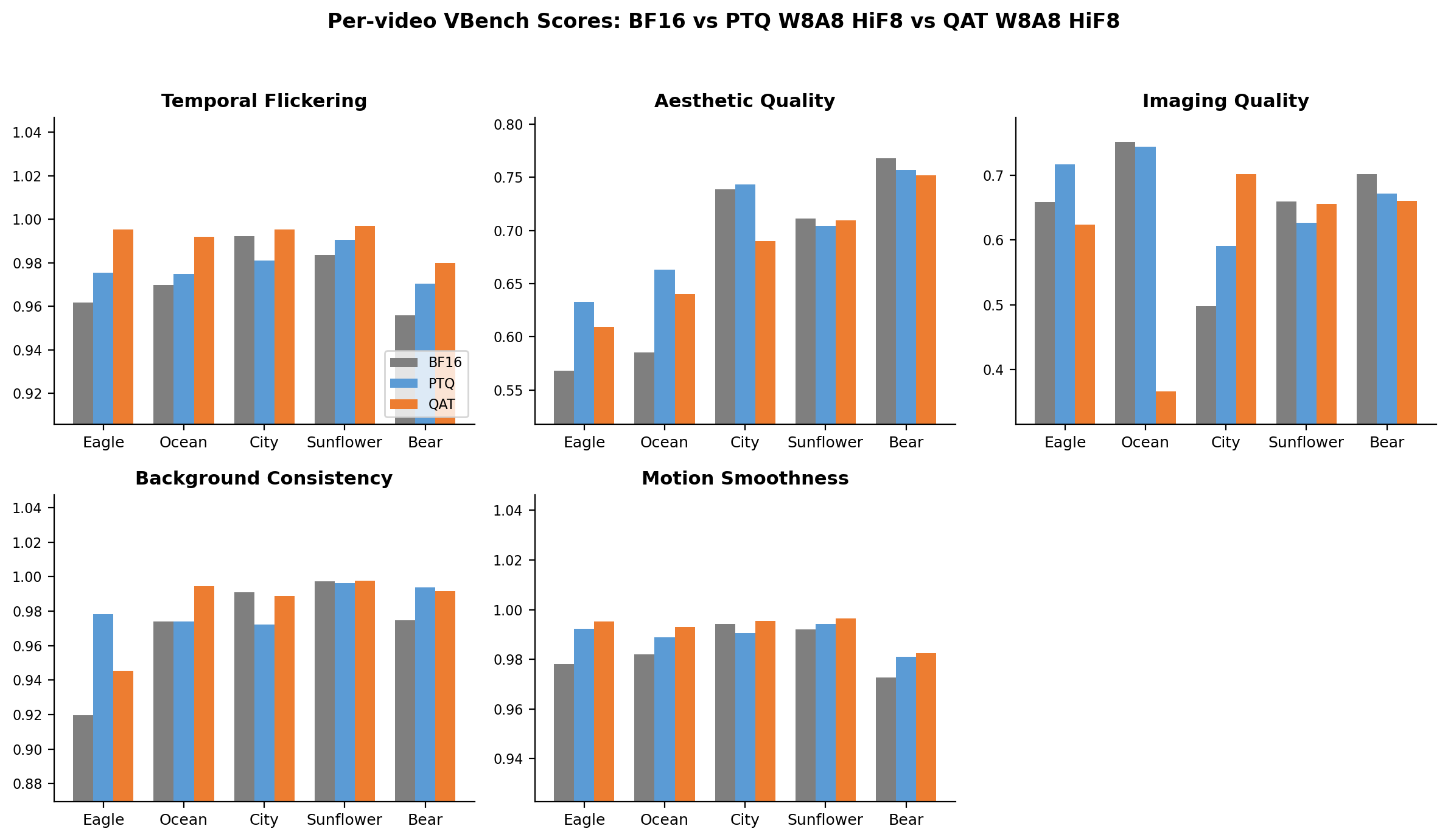}
  \caption{Per-video VBench scores for all three methods. PTQ (blue) matches or exceeds BF16 (gray) on all videos and dimensions. QAT (orange) shows instability on imaging quality (notably video 1, ocean waves).}
  \label{fig:per_video}
\end{figure}

Figure~\ref{fig:per_video} shows the per-video breakdown. PTQ scores are consistently at or above BF16 across all five videos and all five dimensions, confirming that the boundary-protection strategy prevents accuracy degradation rather than merely compensating for it on average.

\subsection{Qualitative Comparison}

\begin{figure*}[t]
  \centering
  \includegraphics[width=\linewidth]{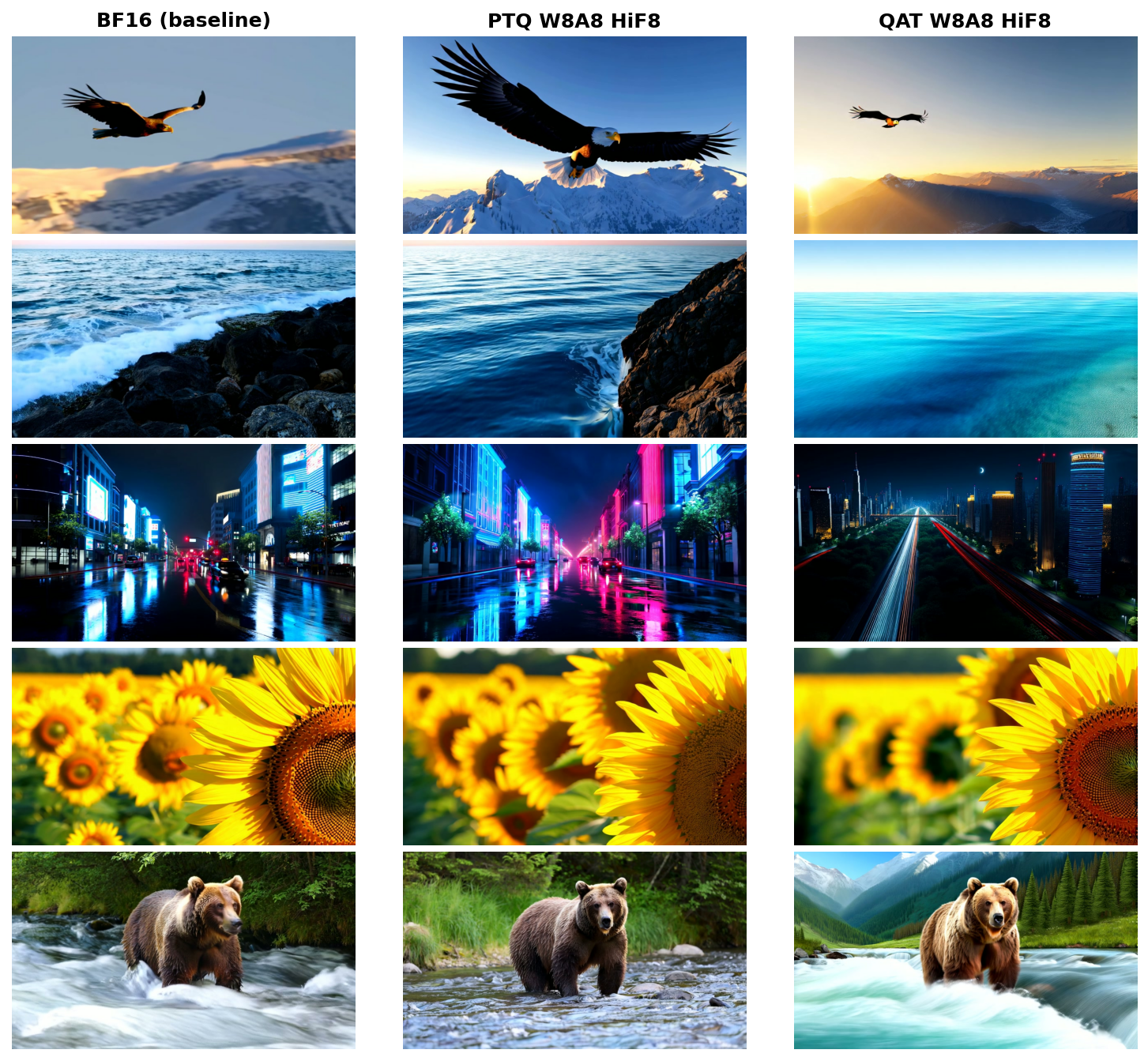}
  \caption{Qualitative comparison across all five prompts (rows) and three methods (columns). Each cell shows frame 24 (mid-sequence) of the generated video. PTQ (center column) closely matches the BF16 baseline (left column) in detail and color fidelity. QAT (right column) exhibits reduced sharpness on several scenes.}
  \label{fig:frames}
\end{figure*}

Figure~\ref{fig:frames} shows mid-sequence frames for all five prompts. PTQ outputs are visually indistinguishable from BF16, preserving fine details (feathers, foam, neon reflections). QAT shows reduced sharpness and altered color balance on several scenes, consistent with its lower IQ scores in Table~\ref{tab:main}.

\subsection{Analysis of QAT Underperformance}

Despite convergent loss, QAT underperforms PTQ on imaging quality ($-0.068$, a 10\% relative drop) due to two factors: (1)~\emph{sequence-length shift}---latents cropped to $\frac{1}{7.4}$ resolution during training cause blocks 2--4 to overfit to short sequences and behave unexpectedly at full 20280-token inference; and (2)~\emph{limited coverage}---only 3 of 35 quantized blocks are fine-tuned, leaving cumulative error in the remaining 32 blocks unaddressed. Both stem from single-card memory constraints, suggesting that reliable video DiT QAT requires multi-card training or aggressive memory-reduction techniques.

\subsection{Ablation Study: Protection Configuration}

To validate the boundary-protection strategy with quantitative evidence, we compare four protection configurations, all using W8A8 HiF8 for non-protected blocks:

\begin{table}[t]
\centering
\caption{Ablation over protection configurations (5-video avg). $\mathcal{P}$: protected block set. Best per-column in \textbf{bold}.}
\label{tab:ablation}
\begin{tabular}{llccccc}
\toprule
Config & $\mathcal{P}$ & TF & AQ & IQ & BC & MS \\
\midrule
None     & $\emptyset$         & 0.970 & 0.651 & 0.663 & 0.971 & 0.983 \\
Entry    & $\{0,1\}$           & 0.971 & 0.655 & 0.655 & 0.972 & 0.984 \\
Exit     & $\{37{-}39\}$       & 0.970 & 0.653 & 0.656 & 0.968 & 0.983 \\
Both     & $\{0,1,37{-}39\}$   & \textbf{0.978} & \textbf{0.700} & \textbf{0.670} & \textbf{0.983} & \textbf{0.989} \\
\bottomrule
\end{tabular}
\end{table}

Table~\ref{tab:ablation} reveals three findings. First, \emph{full boundary protection is essential}: the ``Both'' configuration outperforms all partial configurations by 1.7--1.9 percentage points on average VBench score. Second, \emph{neither entry nor exit protection alone suffices}---each provides only marginal improvement over no protection, confirming that error containment requires addressing both the computation root and the decoder interface simultaneously. Third, the largest gains appear on aesthetic quality (+4.9pp) and background consistency (+1.5pp), dimensions most sensitive to global coherence, suggesting that boundary blocks disproportionately influence scene-level rather than frame-level quality.

\begin{figure*}[t]
  \centering
  \includegraphics[width=\textwidth]{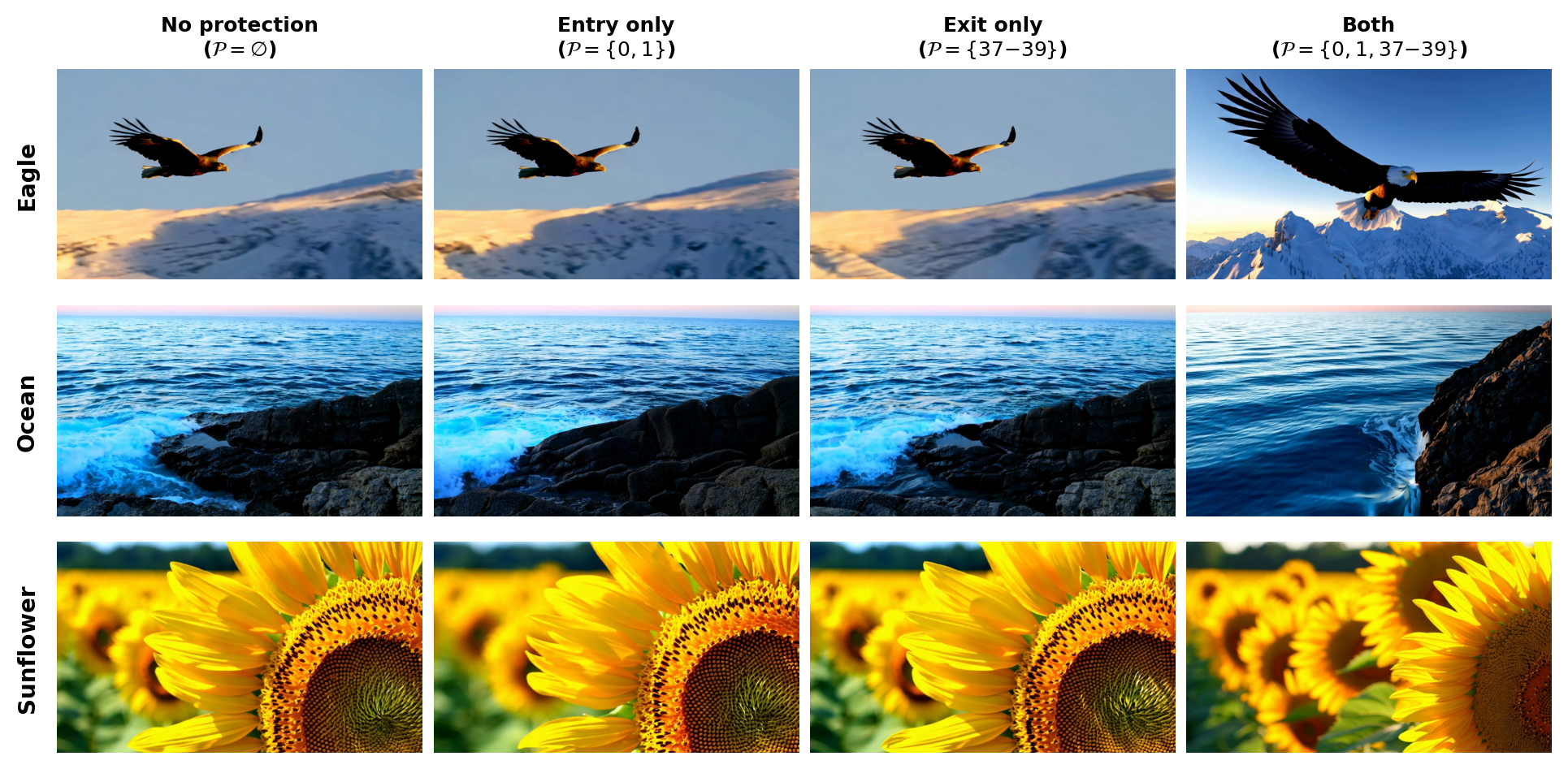}
  \caption{Ablation qualitative comparison (frame 24). Columns: four protection configurations. Full boundary protection (rightmost) produces the sharpest details and most vivid colors.}
  \label{fig:ablation}
\end{figure*}

Figure~\ref{fig:ablation} provides qualitative evidence: the ``Both'' configuration produces sharper details and more vivid colors (e.g., eagle plumage, wave textures), while partial or no protection yields flatter tones and reduced contrast.

\subsection{Deployment on Ascend 910B NPUs}

All experiments run on Ascend 910B NPUs (64\,GB HBM) with \texttt{torch\_npu} and CANN 8.2.RC1. HiF8 is the native 8-bit format of the Ascend Cube compute units, so the W8A8 quantization maps directly to hardware-accelerated matrix multiplication without software simulation. Quantizing 35 of 40 blocks from BF16 to HiF8 saves $\sim$12\,GB of weight memory, keeping inference within single-card HBM. Under model offloading, BF16 generates a 49-frame video in 763\,s (57.2\,GB peak HBM); the PTQ model requires 887\,s (58.7\,GB) due to software quantize--dequantize overhead in the current toolkit---native HiF8 Cube execution eliminates this cost.

\section{Conclusion}

This paper presented a data-driven boundary-protection strategy for W8A8 HiFloat8 quantization of Wan2.1-T2V-14B. By retaining five boundary blocks in BF16 and quantizing the remaining 35 to HiF8, the proposed PTQ method matches or exceeds the BF16 baseline on all five VBench dimensions with zero measurable precision loss. An ablation study confirms that full boundary protection is necessary---neither entry-only nor exit-only protection suffices. QAT under single-card constraints underperforms plain PTQ due to sequence-length shift and limited coverage, highlighting the need for multi-card training infrastructure. A limitation of this study is the small evaluation set (5 prompts), which constrains statistical power; the consistent per-video improvements (Figure~\ref{fig:per_video}) mitigate but do not eliminate this concern, and larger-scale evaluation is needed for definitive conclusions. The boundary-protection principle---preserving error-amplifying entry and high-scale exit blocks---may generalize to other DiT architectures with similar residual-stream topologies.

\section*{Acknowledgment}
Experiments were conducted on Huawei Ascend 910B NPUs provided by the University of Chinese Academy of Sciences. The author thanks the ICME 2026 Grand Challenge organizers and the Global Computing Consortium for providing the HiFloat8 simulation toolkit and evaluation infrastructure.

\end{document}